\documentclass[a4paper,fleqn]{cas-dc}
\usepackage[numbers,sort&compress]{natbib}
\usepackage{algorithmic}
\usepackage{algorithm}
\usepackage{threeparttable}
\usepackage{arydshln}
\usepackage{ulem}
\usepackage{adjustbox}

\ExplSyntaxOn
\keys_set:nn { stm / mktitle } { nologo }
\ExplSyntaxOff

\begin{document}
\let\WriteBookmarks\relax
\def\floatpagepagefraction{1}
\def\textpagefraction{.001}
\shorttitle{}
\shortauthors{}

\title [mode = title]{Demonstration-based learning for few-shot biomedical named entity recognition under machine reading comprehension}

\author[1]{Leilei Su}
\credit{Investigation, Conceptualization, Methodology, Formal analysis, Writing - original draft}

\author[2]{Jian Chen}
\credit{Methodology, Software, Validation}

\author[3]{Yifan Peng}
\credit{Validation, Writing-review \& editing}

\author[2,3]{Cong Sun}[orcid=0000-0003-0240-1070]
\cormark[1]
\ead{cs2565@cornell.edu}
\credit{Supervision, Writing-review \& editing}

\address[1]{Department of Mathematics, Hainan University, Haikou 570228, China}
\address[2]{Department of Data Science and Big Data Technology, Hainan University, Haikou 570228, China}
\address[3]{Department of Population Health Sciences, Weill Cornell Medicine, Cornell University, New York 10065, USA}
\cortext[cor1]{Corresponding author}

\begin{abstract}
Objective: Although deep learning techniques have shown significant achievements, they frequently depend on extensive amounts of hand-labeled data and tend to perform inadequately in few-shot scenarios. The objective of this study is to devise a strategy that can improve the model's capability to recognize biomedical entities in scenarios of few-shot learning.\\
Methods: By redefining biomedical named entity recognition (BioNER) as a machine reading comprehension (MRC) problem, we propose a demonstration-based learning method to address few-shot BioNER, which involves constructing appropriate task demonstrations. In assessing our proposed method, we compared the proposed method with existing advanced methods using six benchmark datasets, including BC4CHEMD, BC5CDR-Chemical, BC5CDR-Disease, NCBI-Disease, BC2GM, and JNLPBA.\\
Results: We examined the models' efficacy by reporting F1 scores from both the 25-shot and 50-shot learning experiments. In 25-shot learning, we observed 1.1\% improvements in the average F1 scores compared to the baseline method, reaching 61.7\%, 84.1\%, 69.1\%, 70.1\%, 50.6\%, and 59.9\% on six datasets, respectively. In 50-shot learning, we further improved the average F1 scores by 1.0\% compared to the baseline method, reaching 73.1\%, 86.8\%, 76.1\%, 75.6\%, 61.7\%, and 65.4\%, respectively.\\
Conclusion: We reported that in the realm of few-shot learning BioNER, MRC-based language models are much more proficient in recognizing biomedical entities compared to the sequence labeling approach. Furthermore, our MRC-language models can compete successfully with fully-supervised learning methodologies that rely heavily on the availability of abundant annotated data. These results highlight possible pathways for future advancements in few-shot BioNER methodologies.
\end{abstract}

\begin{keywords}
few-shot learning\sep prompt-based learning\sep demonstration-based learning\sep machine reading comprehension\sep biomedical named entity recognition
\end{keywords}

\maketitle

\section{Introduction}
Biomedical literature contains a wealth of biomedical entities. Automatic and accurate recognition of these entities, also known as biomedical named entity recognition (BioNER), serves as a fundamental basis for many biomedical research endeavors and plays a significant role in advancing biomedicine.

BioNER aims to automatically identify and classify biomedical entities, such as diseases, chemicals, and proteins, mentioned in unstructured text. Recently, neural networks have shown impressive success, achieving state-of-the-art results on various benchmark datasets~\cite{lample2016neural, ma2016end, liu2018empower}. However, these models, which often require large amounts of manually annotated data, can be expensive and time-consuming to develop, particularly in research domains like biomedicine~\cite{huang2021few,ding2021few}. 

To improve the efficacy of BioNER in low-resource scenarios, previous research has shown a growing interest in leveraging various human-curated resources as auxiliary supervision~\cite{nie2020improving,tian2020improving, wang2021improving, lee2020lean, lin2020triggerner, cui2021template,ding2022prompt, lee2022good}. One such direction is distant-supervised learning, which uses entity dictionaries~\cite{yang2018distantly, shang2018learning, liu2019towards, peng2019distantly} or labeling rules~\cite{safranchik2020weakly,jiang2020cold} to generate noisy-labeled data for training BioNER models. Although utilizing noisy-labeled data can reduce the need for manual annotations, this approach might result in significant amounts of incorrectly labeled data, thereby reducing the model's performance~\cite{meng2021distantly}. 

Another research path uses natural language prompts to boost the performance of natural language processing (NLP) methods in low-resource scenarios~\cite{zhao2021calibrate,brown2020language,yang2020emnlp}. Gao et al. demonstrated that appropriate task demonstrations from the training set can significantly enhance performance~\cite{gao2021making}. Their method concatenates training examples with task demonstrations to enhance the prediction accuracy in cloze-style questions. Further exploration includes optimizing selection of training examples~\cite{gao2021making}, permutating training examples for demonstrations~\cite{kumar2021reordering}, and selecting demonstrations based on the scores from the development set~\cite{lee2022good}. While these methods offer competitive performance, they still face significant challenges, such as the substantial human efforts required to create auxiliary supervision like dictionaries, rules, and explanations, and the high computational costs required for making predictions, especially for long and complex sentences~\cite{lee2022good, ding2022prompt}. With the rise of language models, the MRC framework has been widely used in various NLP tasks, such as question answering~\cite{rajpurkar2016squad}, information retrieval~\cite{fan2022pretraining}, and text summarization~\cite{wu2018learning}. For NER tasks, both Li \cite{li2020mrc} and Sun \cite{sun2021biomedical} provide strong evidence that transforming traditional sequence labeling problems into machine reading comprehension (MRC) tasks not only addresses limitations inherent in sequence labeling models, such as handling overlapping entities and incorporating prior knowledge, but also significantly boosts the overall performance across a wide range of datasets. This shift in paradigm allows models to leverage natural language queries to extract entities, resulting in improved generalization and flexibility in both flat and nested NER tasks.

\begin{figure*}
\centerline{\includegraphics[width=0.8\linewidth]{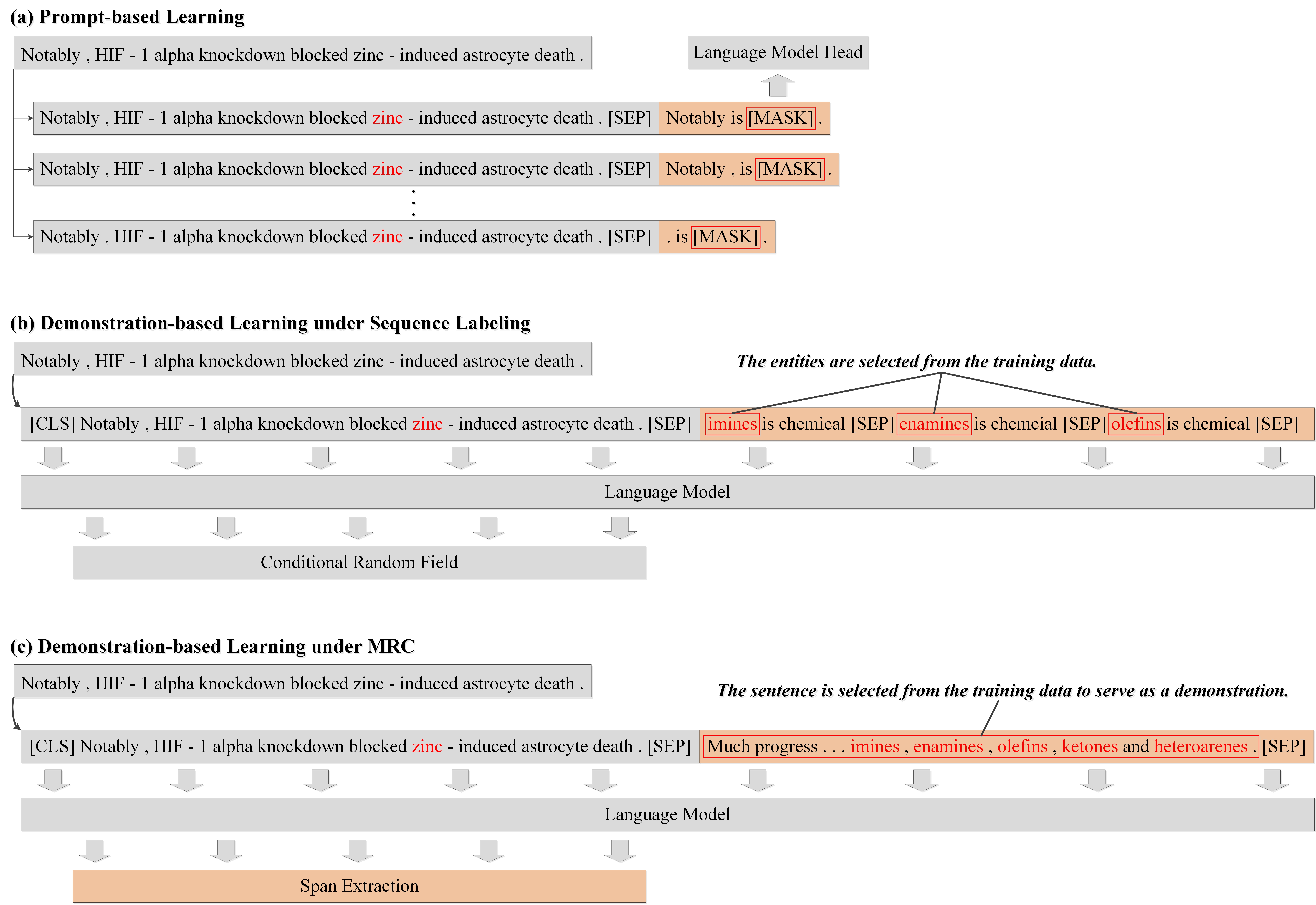}}
\caption{Examples to illustrate the differences between prompt-based learning, demonstration-based learning under sequence labeling, and demonstration-based learning under MRC. (a) Prompt-based learning methods often overlook entity span detection, leading to a lengthy process of generating prompts for all entity candidates. (b) Demonstration-based learning under sequence labeling can truncate the demonstration sequence during training and prediction. (c) Demonstration-based learning under MRC can effectively improve few-shot BioNER performance, provided that suitable task demonstrations are used.}
\label{fig:overview}
\end{figure*}

In this study, we propose a demonstration-based learning method for few-shot BioNER to address these issues. Specifically, we frame the BioNER task as an MRC problem and facilitate in-context learning by constructing appropriate task demonstrations. The MRC framework facilitates extracting answer spans from a given context, guided by a specific query. Inspired by the key ideas in MRC~\cite{sun2021biomedical}, our proposed method breaks down the BioNER task into two classification problems: predicting the start and end position of the answer span within the context. We hypothesize that BioNER methods leveraging the MRC framework could achieve significant improvement in the few-shot settings. 

Our method concatenates elaborately structured task demonstrations to the input sequence and feeds them into the MRC framework. This approach allows the model to better understand the input sequence and learn more nuanced semantic representations. Unlike previous methods that rely on manual annotated auxiliary supervision such as dictionaries, rules, and explanations, our method can automatically select suitable task demonstrations from the training data. In contrast to methods that convert token classification to a cloze-style mask-filling prediction format, our method can seamlessly integrate with pre-trained language models without significant time overhead. Notably, it also facilitates a deeper understanding of contextual relationships between input sentences and task demonstrations compared to methods that truncate the demonstration sequence during both model training and prediction (Figure~\ref{fig:overview} (c) vs. (a) and (b)).

To evaluate the effectiveness of our method in few-shot settings, we conducted extensive experiments across six benchmark datasets. These include BC4CHEMD~\cite{krallinger2015chemdner}, BC5CDR-Chemical~\cite{li2016cdr}, BC5CDR-Disease~\cite{li2016cdr}, NCBI-Disease~\cite{doğan2014ncbi}, BC2GM~\cite{smith2008bc2gm}, and JNLPBA~\cite{collier2004jnlpba}. Our results show that MRC-based language models exhibit significant improvement in BioNER compared to sequence labeling methods. More notably, MRC-based language models can achieve highly competitive performance levels, even when compared to fully supervised learning methods that rely heavily on large amounts of annotated data. The code can be found at \url{github.com/cos4007/grape_demonstrations}.

\section{Materials and Methods}
\subsection{Datasets}
Our method is evaluated on six biomedical datasets, namely BC4CHEMD~\cite{krallinger2015chemdner}, BC5CDR-Chemical (BC5CDR-C)~\cite{li2016cdr}, BC5CDR-Disease (BC5CDR-D)~\cite{li2016cdr}, BC2GM~\cite{smith2008bc2gm}, NCBI-Disease~\cite{doğan2014ncbi}, and JNLPBA~\cite{collier2004jnlpba} (Table \ref{tab:data}). BC5CDR-C is specifically utilized to evaluate chemical entities, while BC5CDR-D is exclusively used to evaluate disease entities. These six datasets are all pre-processed and provided by Lee~\cite{lee2019biobert}. 

To simulate the low-resource setting in real-world situations, we first utilized five random seeds to arbitrarily select 25 and 50 unique sentences from each dataset to be the training sets. Then, we randomly chose 100 sentences from each dataset to serve as development sets. The test sets remain unchanged from the original datasets. Using this approach, we constructed five training sets containing 25 sentences each (Train$_{25}$), five training sets containing 50 sentences each (Train$_{50}$), and five development sets containing 100 sentences each (Dev$_{100}$). 
An experiment conducted on a dataset consisting of Train$_{25}$, Dev$_{100}$ and Test is referred to as a \textbf{25-shot} learning experiment. Similarly, an experiment conducted on a dataset consisting of Train$_{50}$, Dev$_{100}$ and Test is referred to as a \textbf{50-shot} learning experiment, and the same principle applies in other cases. 

\begin{table}[!t]
\caption{The statistics of six benchmark datasets. The numbers represent the average number of entities across five unique samples. Train+Dev refers to the total training and development sets provided by Lee~\cite{lee2019biobert}.\label{tab:data}}
\begin{threeparttable}
\begin{adjustbox}{width=\columnwidth}
\begin{tabular}{lr@{}lr@{}lr@{}lr}
\toprule
Dataset&\multicolumn{2}{c}{Train$_{25}$}&\multicolumn{2}{c}{Train$_{50}$}&\multicolumn{2}{c}{Dev$_{100}$}&Train+Dev\\
\midrule
BC4CHEMD          &44.8&$_{\pm{2.2}}$ &97.4&$_{\pm{9.6}}$ &207.2&$_{\pm{16.6}}$ &58,964\\
BC5CDR-C          &44.6&$_{\pm{2.0}}$ &86.2&$_{\pm{2.3}}$ &179.2&$_{\pm{16.4}}$ &10,550\\
BC5CDR-D          &41.4&$_{\pm{8.3}}$ &78.2&$_{\pm{6.9}}$ &164.8&$_{\pm{6.7}}$  &8,428\\
NCBI-Disease      &40.4&$_{\pm{3.6}}$ &82.2&$_{\pm{4.8}}$ &166.6&$_{\pm{2.4}}$  &5,921\\
BC2GM             &58.4&$_{\pm{8.1}}$ &114.2&$_{\pm{5.4}}$ &227.4&$_{\pm{18.2}}$ &18,258\\
JNLPBA            &65.2&$_{\pm{5.8}}$ &128.6&$_{\pm{5.4}}$ &263.8&$_{\pm{28.1}}$ &40,753\\
\bottomrule
\end{tabular}
\end{adjustbox}
\end{threeparttable}
\end{table}

\subsection{Problem Definition}
Given an input sentence $x = \{x_1, x_2, \cdots, x_n\}$, each word $x_i$ is associated with a corresponding NER label $y_i$, forming a label sequence $y = \{y_1, y_2, \cdots, y_n\}$. Applying NER in the sequence labeling framework aims to predict the entity label $y_i \in Y$ for each word $x_i$, where $Y$ is a pre-defined set of labels combined with IOB2 schema~\cite{sang2000introduction}, which represents the beginning (B) of a phrase, the interior (I) of a phase, and other words (O). For example, in the BC4CHEMD dataset, the sentence ``Much progress has been successfully achieved in the asymmetric reduction of imines, enamines, olefins, ketones, and heteroarenes .'' is labeled as ``O O O O O O O O O O O B-Chemical O B-Chemical O B-Chemical O B-Chemical O B-Chemical O'', where ``imines'', ``enamines'', ``olefins'', ``ketones'', and ``heteroarenes'' are chemicals.

Conversely, when using MRC, the NER task is converted to a set of triples \textit{(Context, Query, Answer)}. \textit{Context} refers to the input sentence $x$, \textit{Query} refers to an auxiliary sentence constructed based on $x$, and \textit{Answer} represents target entity spans. While constructing each biomedical type in the dataset, the Context $x$ is concatenated with a constructed demonstration Query $q = \{q_1, q_2, \cdots, q_m\}$, where $m$ represents the length of the constructed demonstration. As a result, the dataset originally labeled under the sequence labeling framework is transformed into a span-indexed annotation format under the MRC framework. For example, the token-tag pair of the sentence above can be transformed to ``[CLS] Notably, HIF - 1 alpha knockdown blocked zinc - induced astrocyte death. [SEP] Much progress has been successfully achieved in the asymmetric reduction of imines, enamines, olefins, ketones, and heteroarenes. [SEP]'' and ``[CLS] 0 0 0 0 0 0 0 0 1 0 0 0 0 0 [SEP] 0 0 0 0 0 0 0 0 0 0 0 1 0 1 0 1 0 1 0 [SEP]'', where [CLS] and [SEP] are special tokens. Finally, the labeled entities $x_{start, end}$ (e.g., $x_{8,9}$=zinc) can be obtained according to a specific label type (e.g., Chemical). $x_{start, end}$ denotes the continuous tokens from $start$ to $end$ in the sentence $x$, provided that $start < end$.

\subsection{Demonstration Construction}
Let $D_{train}$, $D_{dev}$, and $D_{test}$ be the annotated train, development, and test sets, respectively. We aim to explore the effectiveness of model performance when annotated instances in $D_{train}$ and $D_{dev}$ are extremely limited. 

For language models like BERT, incorporating domain knowledge through demonstrations for auxiliary supervision is highly valuable. Therefore, effectively constructing demonstrations to help these models fully leverage limited annotated data is crucial. Intuitively, the denser the entity in the task demonstration, the increased performance we should observe in a few-shot learning scenario. In this study, we introduce a novel demonstration type termed ``grape'' demonstrations to enhance BioNER performance under the MRC framework. Specifically, we select sentences with the highest entity density from the training set as demonstration sentences. Given their high entity density, we metaphorically refer to them as ``grape'' demonstrations, akin to clusters of grapes on a vine. Below is the density formula used to select grape demonstrations.

\begin{equation}
f_{density} = \frac{(count(entity))^u}{(len(sentence))^v}
\end{equation}
where $count(entity)$ denotes the number of entities and $len(sentence)$ denotes the number of words in the sentence derived from $D_{train}$. The exponents $u$ and $v$ can be any positive values. In the experiments, these two values were empirically set.

We determine the density rankings based on $f_{density}$ and select the sentence with the highest density as the demonstration. However, if the selected demonstration contains too many words, the combined sequence of $x$ and $q$ could exceed the maximum length allowed in the language model. Therefore, we impose a limit on the length of the demonstration; if it exceeds the limit, we discard the demonstration.

\subsection{Model Details}
Given a demonstration $q$, we aim to extract the continuous text span $x_{start, end}$ from the input sentence $x$. We use BERT~\cite{devlin2019bert} as the backbone and employ PubmedBERT~\cite{gu2021pubmedbert}, a language model specifically pre-trained on biomedical texts, to improve the method's performance. We concatenate $x$ and $q$ into a combined sequence $\{\text{[CLS]}\: x\: \text{[SEP]}\: q\: \text{[SEP]}\}$, where the length of the combined sequence is $n$. Then, we feed the combined sequence into the BERT model. Finally, the BERT model generates a contextual representation matrix $H \in \mathbb{R}^{n \times d}$, where $d$ represents the vector dimension of the last layer of the BERT model. During inference, the [CLS], [SEP], and demonstration representations are dropped.

In this study, we construct two separate binary classifiers to predict a target entity's start and end positions for the span selection. This dual classifier approach allows for the output of multiple start and end indexes for a single input sequence, and therefore it has the potential to identify all the desired entities. 

Our method begins by predicting the probability of each token being a start index. 
\begin{equation}
P_{start} = linear(H\cdot W_{start}) \in \mathbb{R}^{n \times 2}
\end{equation}
where $W_{start} \in \mathbb{R}^{d \times 2}$ represents the learned weights and $linear(\cdot)$ denotes a fully connected layer. Each row of the resulting $P_{start}$ matrix represents the probability distribution of each index within the input sequence serving as a potential starting position for an entity.

Then, our method proceeds to predict the probability of each token being the corresponding end index. 
\begin{equation}
P_{end} = linear(H\cdot W_{end} \oplus \text{softmax}(P_{start})) \in \mathbb{R}^{n \times 2}
\end{equation}
where $W_{end} \in \mathbb{R}^{d \times 2}$ represents the learned weights and $\oplus$ is the concatenation operation. The resulting $P_{end}$ matrix contains the probability distribution of each index serving as a potential ending position for an entity.

Finally, we obtain the predicted start and end positions for potential entities by applying the $\arg\max$ function to each row of $P_{start}$ and $P_{end}$.
\begin{align}
I_{start} &= \{i | \arg\max(P^i_{start}) = 1, i = 1, 2, \cdots, n\}\\
I_{end} &= \{j | \arg\max(P^j_{end}) = 1, j = 1, 2, \cdots, n\}
\end{align}
where the superscript $i$ and $j$ denote the $i$-th and $j$-th rows of a matrix, respectively.

In the input sentence $x$, it is possible to have multiple entities of the same type, which may result in predicting multiple start and end indexes. Since the BioNER datasets in this study do not comprise overlapping entities, we use the unidirectional nearest match principle to obtain the final predicted answer spans. Specifically, in cases where one end/start index matches multiple start/end indexes, we simply pair it with the closest start/end index and discard others, following the approach by Sun \cite{sun2021biomedical}.

\subsection{Training and Inference}
During the training process, we assign two two label sequences $y_{start}$ and $y_{end}$ for each input sentence $x$. 
The loss function is then defined as:
\begin{align}
L_{start} &= \text{cross-entropy}(P_{start}, y_{start})\\
L_{end} &= \text{cross-entropy}(P_{end}, y_{end})\\
L &= (L_{start}+L_{end})/2
\end{align}

During the inference phase, the start and end indexes are independently determined using $I_{start}$ and $I_{end}$, respectively. Subsequently, the unidirectional nearest match principle is applied to align the start and end indexes, producing the final output.

\section{Experiments}
\subsection{Experimental Setup}
To demonstrate the effectiveness of our method, we constructed two baseline methods: BERT-CRF$_{vanilla}$ and BERT-CRF$_{popular}$. BERT-CRF$_{vanilla}$ is a traditional sequence labeling model that combines BERT and CRF. It employs BERT as the encoder to convert each word in the input sequence into its corresponding representation. Then, it utilizes CRF to predict label sequences. BERT-CRF$_{popular}$ combines the original sentence with the popular type demonstration, creating a composite input for BERT-CRF. The popular type, as proposed by Lee et al.~\cite{lee2022good}, is an effective and efficient demonstration type, which selects the sentence containing the entity with the highest occurrence as the demonstration.

In this study, we introduce the MRC framework and grape demonstrations to improve the model's BioNER capabilities. BERT-CRF$_{grape}$ integrates the grape demonstrations into the original sentence to form the input of BERT-CRF. BERT-MRC$_{popular}$ combines the original sentence with the popular-type demonstration to create the input of BERT-MRC. BERT-MRC$_{grape}$ concatenates the original sentence with the grape demonstrations to construct the input. 

For the construction of popular demonstration, we first gather all annotated entities in the training set $D_{train}$ and rank each entity by its frequency of occurrence in $D_{train}$. Then, we select the sentence containing the entity that appears most frequently in $D_{train}$ as the popular demonstration. For constructing the grape demonstration, we set $u$=3, $v$=1 and aim for a maximum demonstration length of 100 to achieve high-quality demonstrations. Next, we calculate the entity density for each sentence in $D_{train}$ using formula (1), and select the optimal demonstration as the grape demonstration.

We implemented all baseline models and our method using PyTorch~\cite{paszke2019nips} and HuggingFace~\cite{wolf2020emnlp}. In the experiments, PubmedBERT$_{large}$ was employed as the weights for the BERT model, and the batch size and learning rate were set to 1 and 3e-5, respectively. Five runs were conducted for each variant with their respective five random seeds on the corresponding datasets, and the average and standard deviation of F1 scores were reported. The F1 score is used to evaluate the model's performance for all methods.
\begin{align}
F1 &= 2PR/(P+R)\\
P &= T\!P/(T\!P+F\!P)\\
R &= T\!P/(T\!P+F\!N)
\end{align}
where $P$, $R$, $T\!P$, $F\!P$, and $F\!N$ denote the precision, recall, true positives, false positives, and false negatives, respectively.

\subsection{Results}
\subsubsection{BERT-CRF vs. BERT-MRC}
Table~\ref{tab:comparison} shows the performance comparison in 25-shot learning and 50-shot learning scenarios. We observed that neural network models have achieved competitive model performance. For BERT-CRF, both BERT-CRF$_{popular}$ and BERT-CRF$_{grape}$ demonstrated a notable performance improvement compared to BERT-CRF$_{vanilla}$. This demonstrates the effectiveness of incorporating demonstrations into the models. When comparing BERT-MRC and BERT-CRF, BERT-MRC shows a more significant performance improvement. This suggests that the MRC framework can greatly enhance the method's performance in low-resource scenarios.

\begin{table}[!tbp]
\caption{Performance comparison in the average and standard deviation of F1 scores (\%) over five runs. 25 and 50 denote 25-shot learning and 50-shot learning, respectively.\label{tab:comparison}}
\centering
\begin{threeparttable}
\begin{adjustbox}{width=\columnwidth}
\begin{tabular}{l *{3}{cc}}
\toprule
\multirow{2}{*}{Method}&\multicolumn{2}{c}{BC4CHEMD} &\multicolumn{2}{c}{BC5CDR-C}\\
\cmidrule(rl){2-3}\cmidrule(rl){4-5}
&25&50&25&50\\
\midrule
BERT-CRF$_{vanilla}$     &53.1$_{\pm{2.8}}$ &66.2$_{\pm{3.6}}$&77.9$_{\pm{3.5}}$&81.7$_{\pm{2.3}}$\\
BERT-CRF$_{popular}$ &56.2$_{\pm{1.9}}$ &67.3$_{\pm{4.0}}$ &78.0$_{\pm{2.6}}$ &83.1$_{\pm{0.7}}$  \\
BERT-CRF$_{grape}$   &57.1$_{\pm{3.8}}$ &67.4$_{\pm{3.1}}$ &78.5$_{\pm{2.5}}$ &82.0$_{\pm{1.6}}$  \\
BERT-MRC$_{popular}$   &61.2$_{\pm{3.4}}$ &71.6$_{\pm{2.6}}$ &83.6$_{\pm{1.2}}$ &86.7$_{\pm{0.7}}$  \\
BERT-MRC$_{grape}$   &\textbf{61.7}$_{\pm{2.1}}$ &\textbf{73.1}$_{\pm{3.1}}$ &\textbf{84.1}$_{\pm{1.4}}$ &\textbf{86.8}$_{\pm{0.6}}$    \\
\midrule
\multirow{2}{*}{Method}&\multicolumn{2}{c}{BC5CDR-D}&\multicolumn{2}{c}{NCBI-Disease}\\
\cmidrule(rl){2-3}\cmidrule(rl){4-5}
&25&50&25&50\\
\midrule
BERT-CRF$_{vanilla}$   &65.6$_{\pm{3.4}}$&71.9$_{\pm{3.8}}$&61.2$_{\pm{4.6}}$&71.7$_{\pm{0.9}}$  \\
BERT-CRF$_{popular}$   &67.4$_{\pm{3.2}}$ &73.3$_{\pm{0.7}}$&65.9$_{\pm{4.2}}$ &73.0$_{\pm{1.0}}$ \\
BERT-CRF$_{grape}$     &64.0$_{\pm{5.8}}$ &74.7$_{\pm{1.0}}$&64.6$_{\pm{1.8}}$ &72.7$_{\pm{2.8}}$ \\
BERT-MRC$_{popular}$   &\textbf{69.4}$_{\pm{3.1}}$ &75.6$_{\pm{0.9}}$&\textbf{71.2}$_{\pm{0.8}}$ &\textbf{75.8}$_{\pm{2.1}}$ \\
BERT-MRC$_{grape}$     &69.1$_{\pm{2.8}}$  &\textbf{76.1}$_{\pm{1.8}}$&70.1$_{\pm{1.5}}$  &75.6$_{\pm{2.7}}$  \\
\midrule
\multirow{2}{*}{Method}&\multicolumn{2}{c}{BC2GM}&\multicolumn{2}{c}{JNLPBA}\\
\cmidrule(rl){2-3}\cmidrule(rl){4-5}
&25&50&25&50\\
\midrule
BERT-CRF$_{vanilla}$     &48.7$_{\pm{3.8}}$&57.1$_{\pm{1.9}}$&57.3$_{\pm{1.6}}$&63.3$_{\pm{2.1}}$\\
BERT-CRF$_{popular}$ &50.5$_{\pm{2.4}}$ &59.6$_{\pm{2.0}}$ &58.1$_{\pm{2.5}}$ &64.5$_{\pm{0.6}}$\\
BERT-CRF$_{grape}$   &\textbf{51.0}$_{\pm{3.2}}$ &59.7$_{\pm{1.4}}$ &58.6$_{\pm{3.1}}$ &63.0$_{\pm{1.5}}$ \\
BERT-MRC$_{popular}$   &50.0$_{\pm{4.2}}$ &\textbf{61.8}$_{\pm{2.5}}$ &59.1$_{\pm{1.2}}$ &65.1$_{\pm{1.3}}$ \\
BERT-MRC$_{grape}$  &50.6$_{\pm{4.7}}$  &61.7$_{\pm{1.6}}$  &\textbf{59.9}$_{\pm{1.6}}$ &\textbf{65.4}$_{\pm{1.4}}$  \\
\bottomrule
\end{tabular}
\end{adjustbox}
\end{threeparttable}
\end{table}

\subsubsection{Grape vs. Popular Demonstrations}
We also compared the performance under BERT-CRF and BERT-MRC between two types of demonstrations: popular and grape. 
For 25-shot learning, we observed BERT-MRC$_{grape}$ and BERT-CRF$_{grape}$ outperformed their respective counterparts, BERT-MRC$_{popular}$ and BERT-CRF$_{popular}$, on BC4CHEMD, BC5CDR-C, BC2GM, JNLPBA datasets. This suggests that constructing grape demonstrations can improve the model's performance for chemical and protein entities. In contrast, BERT-MRC$_{popular}$ and BERT-CRF$_{popular}$ outperformed their corresponding counterparts BERT-MRC$_{grape}$ and BERT-CRF$_{grape}$ on BC5CDR-D and NCBI-Disease datasets. This implies that, for disease entities, utilizing the popular demonstration is an effective choice in the 25-few learning scenario. 
For 50-shot learning, the use of grape demonstration outperforms popular demonstration on four out of six datasets (except for BC2GM and NCBI-Disease).
Taken together, methods using grape demonstrations showed considerably more favorable results  compared to ones using popular demonstrations.

\section{Discussion}
\subsection{Few-shot vs. Fully Supervised Learning}

\begin{figure}[!t]
\centerline{\includegraphics[width=0.95\columnwidth]{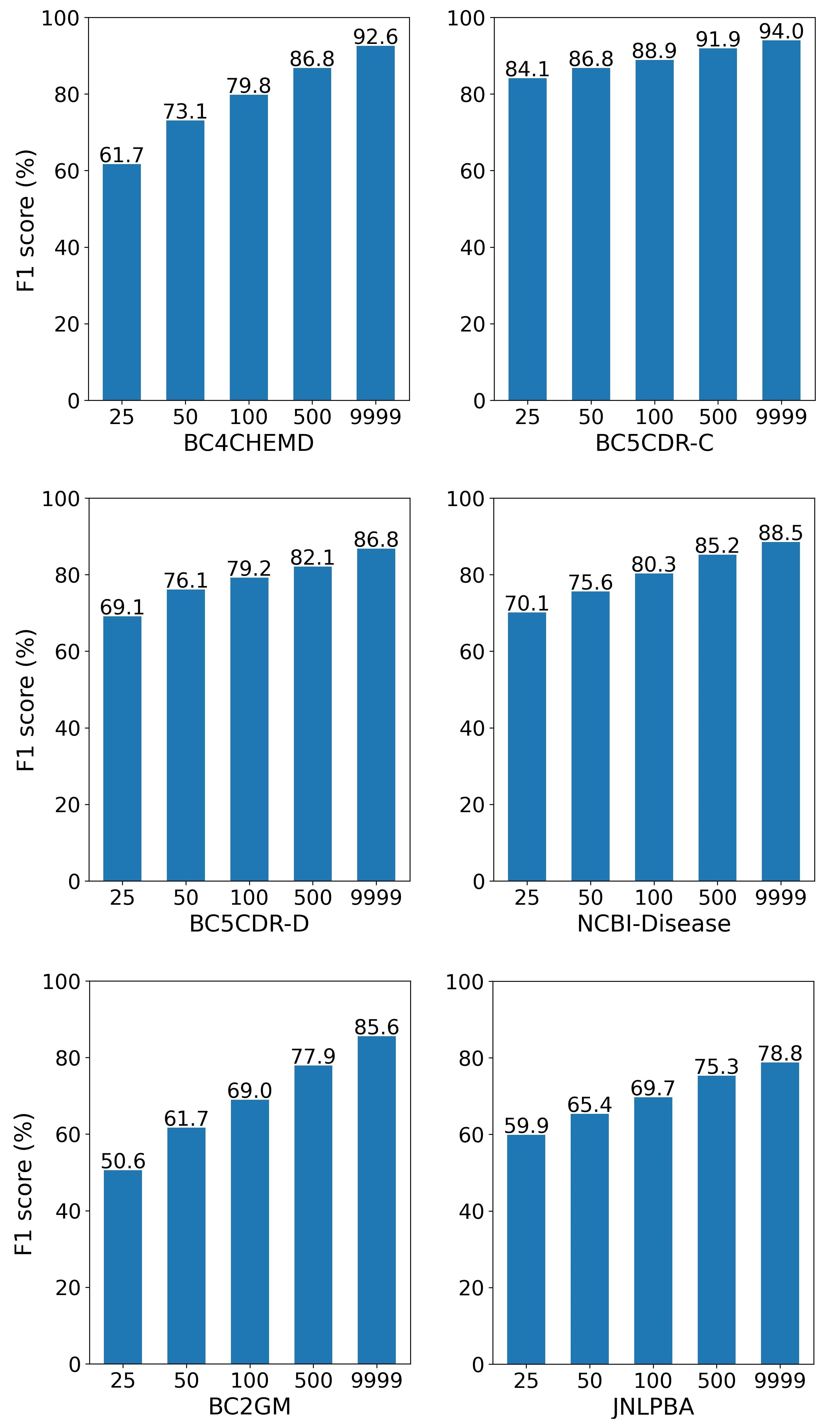}}
\caption{Performance of BERT-MRC$_{grape}$ in 25-shot, 50-shot, 100-shot, 500-shot, and fully supervised (9999) learning. The values are the average F1 scores over five runs.\label{fig:bertmrc}}
\end{figure}

To assess the impact of few-shot on BERT-MRC$_{grape}$, we further compared the model's performance in different ``shots'' (Figure \ref{fig:bertmrc}).
BERT-MRC$_{grape}$ in 25-shot learning obtains F1 scores of 61.7\%, 84.1\%, 69.1\%, 70.1\%, 50.6\%, and 59.9\% on the BC4CHEMD, BC5CDR-C, BC5CDR-D, NCBI-Disease, BC2GM, and JNLPBA datasets, respectively. These F1 scores reach 66.6\%, 89.5\%, 79.6\%, 79.2\%, 59.1\%, and 76.0\% in fully supervised settings.
As the number of samples in the training set increases, the model's performance shows a positive correlation improvement.
In 50-shot learning and 100-shot learning, the corresponding two sets of F1 scores improve to 78.9\%, 92.3\%, 87.7\%, 85.4\%, 72.1\%, and 83.4\%, as well as 86.20\%, 94.60\%, 91.20\%, 90.70\%, 80.60\%, and 88.50\%, respectively, compared to fully supervised settings. 
Finally, the F1 scores in 500-shot learning reach a maximum of 93.7\%, 97.8\%, 94.6\%, 96.3\%, 91.0\%, and 95.6\% in fully supervised settings.

\begin{figure}[!t]
\centerline{\includegraphics[width=0.95\columnwidth]{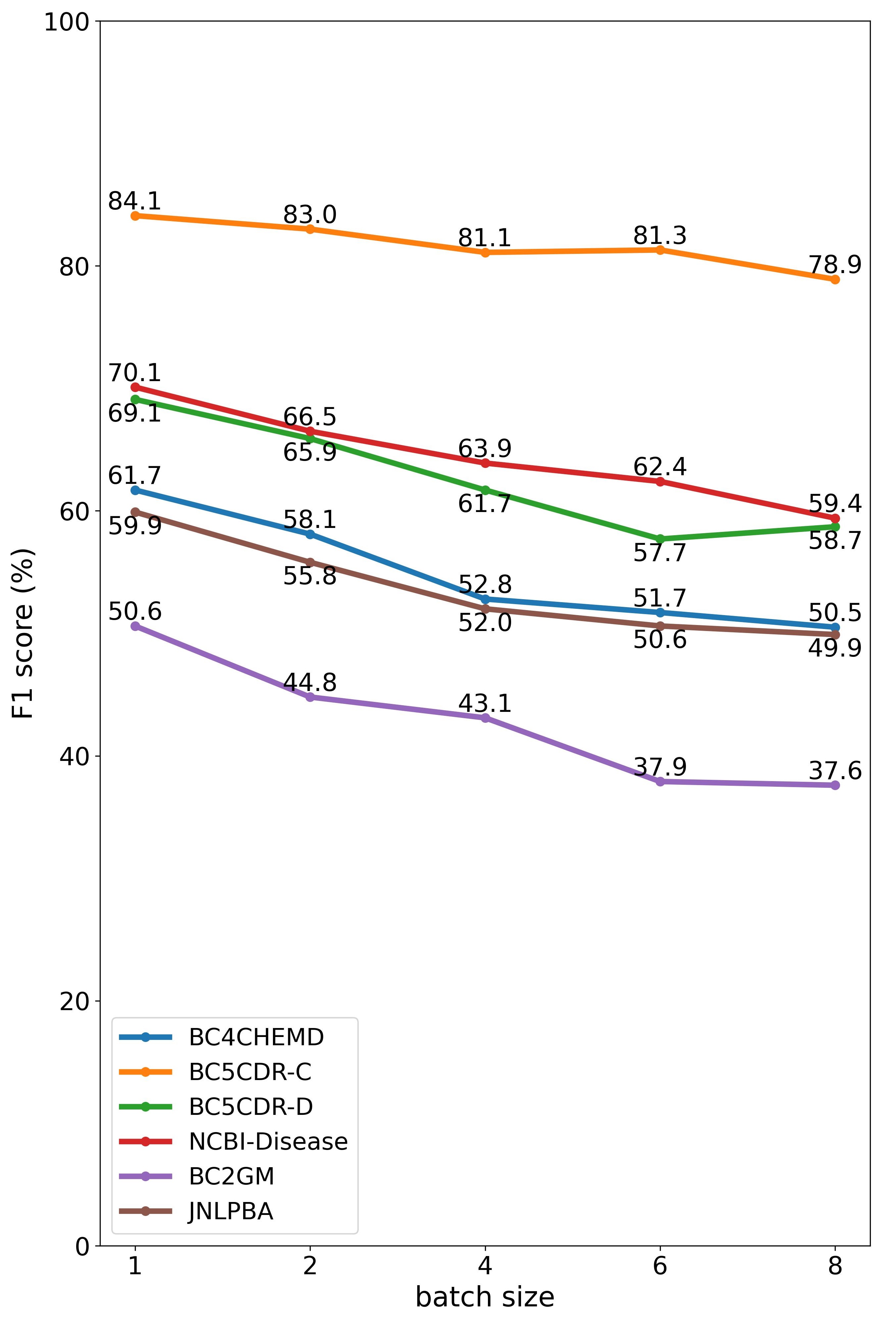}}
\caption{The impact of different batch sizes on the performance of BERT-MRC$_{grape}$ in few-shot learning. The values are the average of F1 scores over five runs.\label{fig:batchsize}}
\end{figure}

It is noteworthy that each dataset used in 25-shot, 50-shot, 100-shot, and 500-shot learning requires only 125, 150, 200, and 600 annotated sentences, respectively. Alternatively, fully supervised settings require 61,321, 9,141, 9,141, 6,347, 15,093, and 18,546 annotated sentences on BC4CHEMD, BC5CDR-C, BC5CDR-D, NCBI-Disease, BC2GM, and JNLPBA, respectively. These experimental results suggest that few-shot learning can leverage a small amount of annotated data to achieve competitive model performance.

\subsection{Batch Size for Few-shot Learning}
Since batch size may greatly impact models' performance, we further investigated the impact of batch size (1, 2, 4, 6, and 8) for training 25-shot learning on BERT-MRC$_{grape}$.
Figure \ref{fig:batchsize} indicates a noticeable variation in F1 values of BERT-MRC$_{grape}$ across different batch sizes in the 25-shot learning setting.  
Overall, for the six BioNER datasets we studied, the F1 scores of BERT-MRC$_{grape}$ consistently decreased as the batch size increased. This occurrence could possibly be attributed to the inherent nature of few-shot learning where, due to the limited number of training samples, it becomes challenging to fully utilize the data when the batch size is increased. Therefore, when setting the batch size to 1, the model can make optimal use of the limited annotated data available for few-shot learning. 

\begin{figure}[!t]
\centerline{\includegraphics[width=0.99\columnwidth]{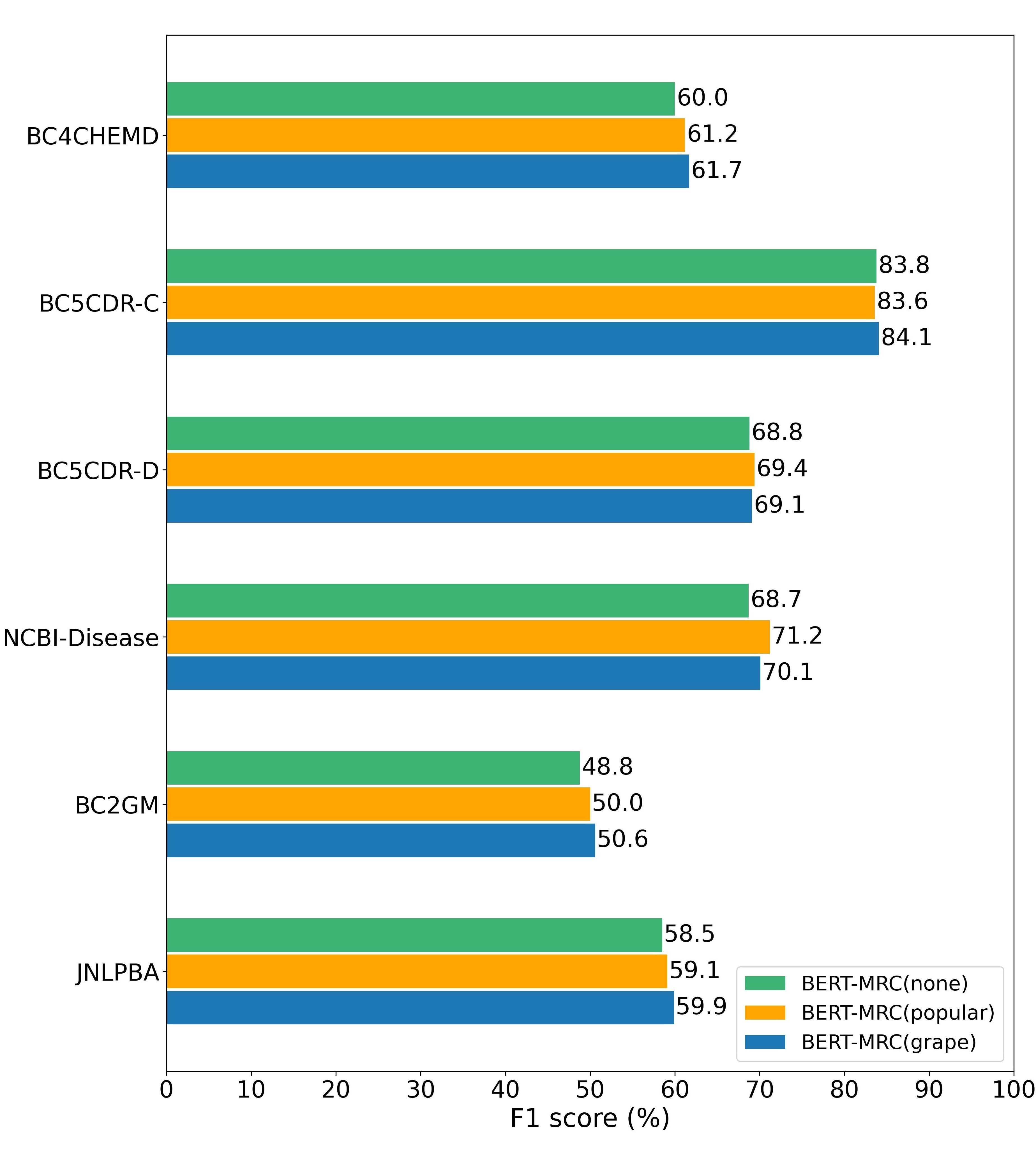}}
\caption{The impact of different demonstrations on the performance of BERT-MRC in few-shot learning. The values are the average of F1 scores over five runs.}
\end{figure}

\subsection{Demonstrations for Few-shot Learning}
We further explored the impact of demonstrations on model performance in few-shot learning. Figure 4 shows the average F1 scores of BERT-MRC$_{none}$, BERT-MRC$_{popular}$, and BERT-MRC$_{grape}$ in 25-shot learning. The average F1 scores of BERT-MRC$_{grape}$ and BERT-MRC$_{popular}$, which utilize meticulously designed demonstrations, significantly surpass that without using demonstrations (BERT-MRC$_{none}$). This suggests that appropriate demonstrations can improve model performance in few-shot learning. Between BERT-MRC$_{popular}$ and BERT-MRC$_{grape}$, BERT-MRC$_{grape}$ achieved superior average F1 scores than BERT-MRC$_{popular}$ on the BC4CHEMD, BC5CDR-C, BC2GM, and JNLPBA datasets and secured competitive average F1 scores on the BC5CDR-D and NCBI-Disease datasets. These results validate the effectiveness of the grape demonstrations proposed in this study.

\begin{figure}[!t]
\centerline{\includegraphics[width=0.8\columnwidth]{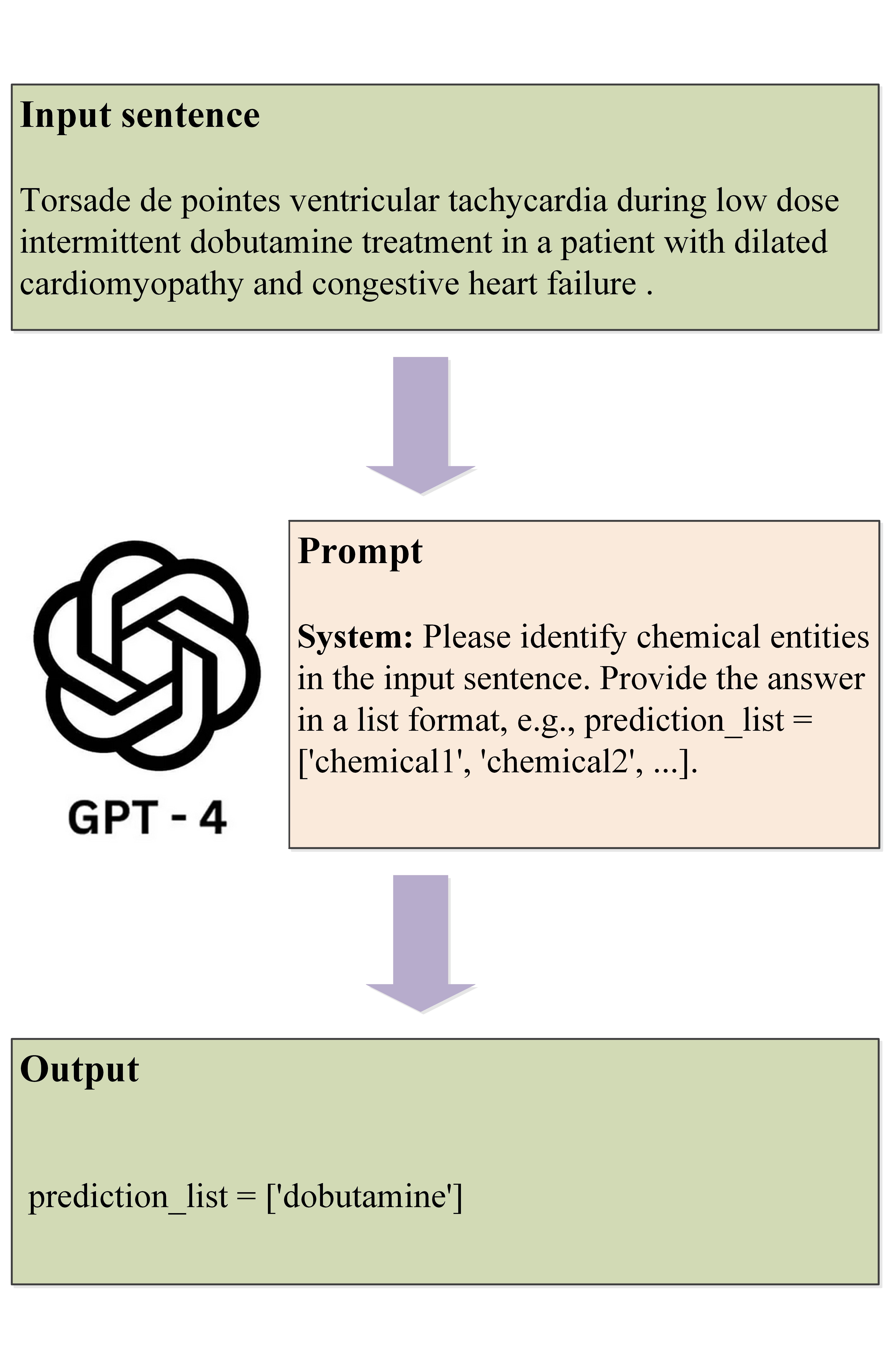}}
\caption{Implementing the BioNER task using GPT-4.}
\end{figure}

\subsection{Demonstration-based Methods vs. GPT-4}
Recently, with the rise of large language models like GPT-4~\cite{openai2023gpt}, there have been new developments in BioNER tasks. In this subsection, we compared our proposed method with GPT-4 to observe the characteristics of these two types of methods. Figure 5 illustrates the process of implementing the BioNER task using GPT-4. In the experiments, we used GPT-4 via Azure OpenAI, with the model version 0125-preview. We designed prompts to instruct GPT-4 to provide target entities based on different input sentences. The prompt we used is: ``Please identify $x$ entities in the input sentence. Provide the answer in a list format, e.g., prediction\_list = [`$x$1', `$x$2', $\dots$].'' Here, $x$ can be replaced with ``chemical'', ``protein'', or ``disease'' depending on the task when using prompts.

\begin{figure}[!t]
\centerline{\includegraphics[width=0.95\columnwidth]{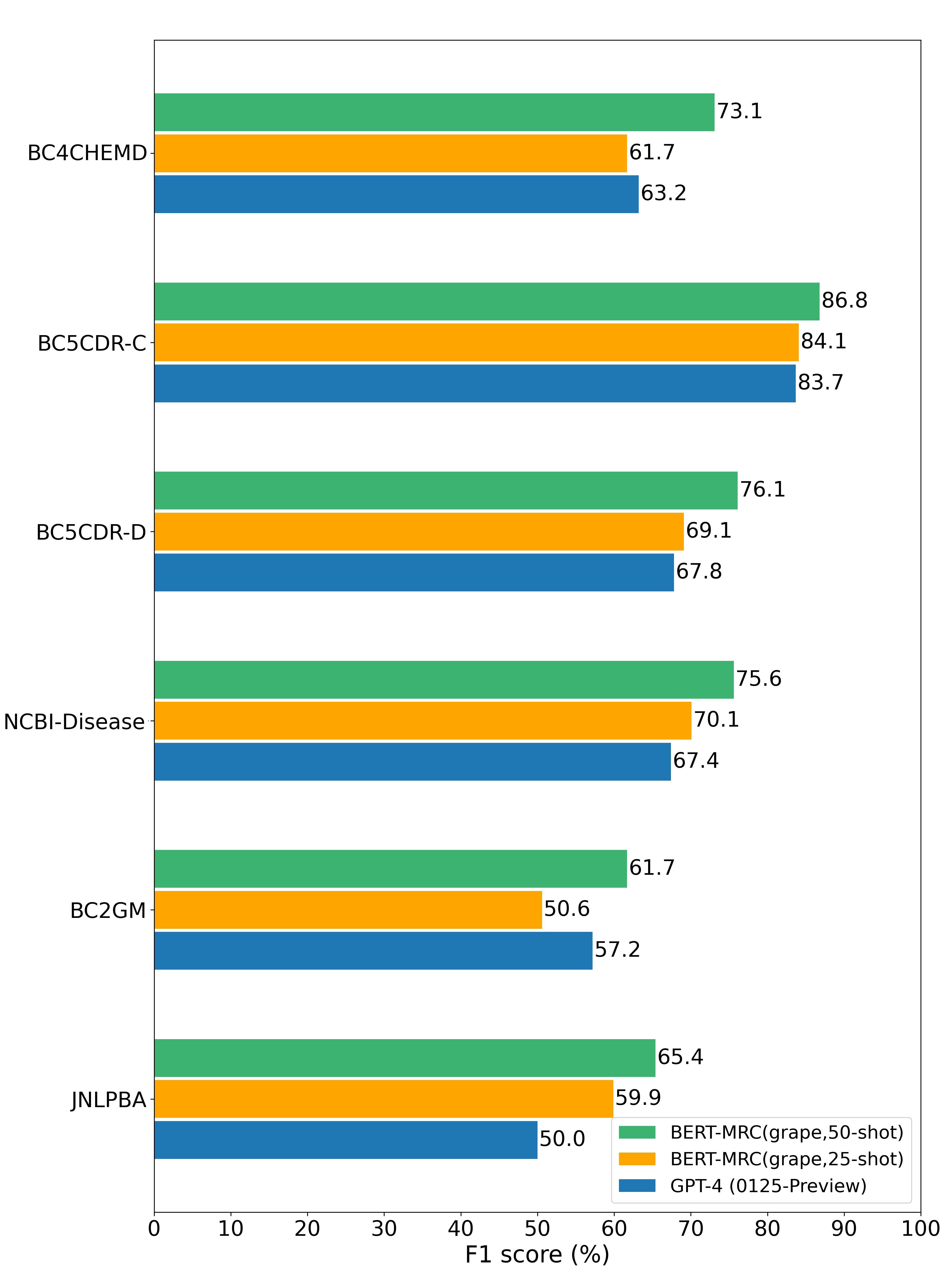}}
\caption{The comparison between our method and GPT-4 on F1 score.}
\end{figure}

Figure 6 shows the comparison between BERT-MRC$_{grape}$ and GPT-4 on F1 scores. It can be seen that under fine-tuning with few-shots, BERT-MRC$_{grape}$ achieves higher F1 scores than GPT-4. In the scenario of 25-shot learning, BERT-MRC$_{grape}$ outperforms GPT-4 across four datasets, while in the scenario of 50-shot learning, BERT-MRC$_{grape}$ consistently outperforms GPT-4. Furthermore, considering the extensive parameters and usage conditions of GPT-4, BERT-MRC$_{grape}$ is also applicable in a broader range of situations, especially in scenarios with limited annotated data.

\section{Conclusion}
In this study, we propose an effective demonstration-based learning method for few-shot BioNER. We reformulated the task as an MRC problem and constructed task demonstrations using a carefully designed density function. The experimental results indicate that the MRC framework is more suitable for BioNER in few-shot learning compared to the CRF framework. The grape demonstration, which we proposed, is a category of effective demonstration that can boost model performance in low-resource scenarios. Our results also show that, under the BERT-MRC framework, few-shot learning exhibits preliminary competitiveness even when compared to fully supervised learning. Moreover, the batch size emerges as a critical hyperparameter. Setting it to 1 in extremely low-resource scenarios can significantly enhance model performance. The reasonable and effective utilization of few-shot learning could have profound implications for assisted healthcare.

\section*{Acknowledgment}
We thank the anonymous reviewers for their constructive comments.

\printcredits
\balance
\bibliographystyle{elsarticle-num}
\bibliography{cas-dc}

\end{document}